\documentclass{article}

\usepackage{arxiv}

\usepackage[utf8]{inputenc} 
\usepackage[T1]{fontenc}    
\usepackage{url}            
\usepackage{booktabs}       
\usepackage{amsfonts}       
\usepackage{nicefrac}       
\usepackage{microtype}      
\usepackage{lipsum}		
\usepackage{doi}
\usepackage{tikz}
\usepackage{amssymb}
\usepackage{float}
\usepackage{subcaption} 
\usepackage{tabularx,ragged2e}

\title{How to Augment for Atmospheric Turbulence Effects on Thermal Adapted Object Detection Models?}

\author{ Engin Uzun \\
Dept. of Image Processing \& Computer Vision Tech. \\
	ASELSAN Inc., Türkiye\\
     Graduate School of Informatics, METU, Türkiye \\
	\texttt{enginuzun@aselsan.com} \\
	\And
    Erdem Akagündüz \\
    Department of Modeling and Simulation \\
    Graduate School of Informatics, METU, Türkiye \\
	\texttt{akaerdem@metu.edu.tr} \\
}

\date{}

\renewcommand{\headeright}{}
\renewcommand{\shorttitle}{How to Augment for Atmospheric Turbulence Effects on Thermal Adapted Object Detection Models?}

\begin{document}
\maketitle
\begin{abstract} 
Atmospheric turbulence poses a significant challenge to the performance of object detection models. Turbulence causes distortions, blurring, and noise in images by bending and scattering light rays due to variations in the refractive index of air. This results in non-rigid geometric distortions and temporal fluctuations in the electromagnetic radiation received by optical systems. This paper explores the effectiveness of turbulence image augmentation techniques in improving the accuracy and robustness of thermal-adapted and deep learning-based object detection models under atmospheric turbulence. Three distinct approximation-based turbulence simulators (geometric, Zernike-based, and P2S) are employed to generate turbulent training and test datasets. The performance of three state-of-the-art deep learning-based object detection models: RTMDet-x, DINO-4scale, and YOLOv8-x, is employed on these turbulent datasets with and without turbulence augmentation during training. The results demonstrate that utilizing turbulence-specific augmentations during model training can significantly improve detection accuracy and robustness against distorted turbulent images. Turbulence augmentation enhances performance even for a non-turbulent test set.
\end{abstract}

\include{bm}



\section{Introduction}
\label{sec:introduction}
A major research area in deep learning and computer vision is augmenting datasets artificially. The philosophy behind data augmentation is to make the training distribution more diverse so that the model is more robust to variations in the input and less prone to overfitting. By creating fresh samples from existing ones through different techniques, the model is exposed to a wider range of variations and can learn to recognize features and patterns that are invariant to various transformations. This helps the model generalize better to unseen data and compensate for imbalanced datasets by generating new samples for under-represented classes and features. Some problem definitions attribute these transformations to complex conditions, such as atmospheric turbulence. 
For a proper introduction to the subject, in the following, we review data augmentation literature before moving on to our problem definition, namely augmenting turbulence effects into thermal-adapted object detection models. 

\subsection{Data Augmentation Literature}

The demand for more labelled data has increased as a result of recent advances in deep learning. Particularly as a result of the explosive growth of vision transformer architectures, attention-based algorithms dominate many vision benchmarks such as image classification \cite{vit2020,wei2022contrastive,wang2022image}, object detection \cite{zhang2022dino, li2022exploring,wang2022image} and image generation \cite{ramesh2022hierarchical,saharia2022photorealistic}. There is, however, significant consumption of labeled data in these algorithms. 
To address this requirement, current deep learning research predominantly focus on utilizing self-supervised \cite{jaiswal2020survey} or semi-supervised learning \cite{zhai2019s4l,yang2021survey} methods. No matter what supervision strategy is used in these studies, training deep neural networks requires more data than ever before. Augmentation is still one of the most efficient and effective ways to increase the amount of labeled or unlabelled data to regularize problems such as small-scale sets, class imbalance, and domain shift, to name a few. 

\begin{figure}[t]
    \centering
    \includegraphics [width=8.5 cm]{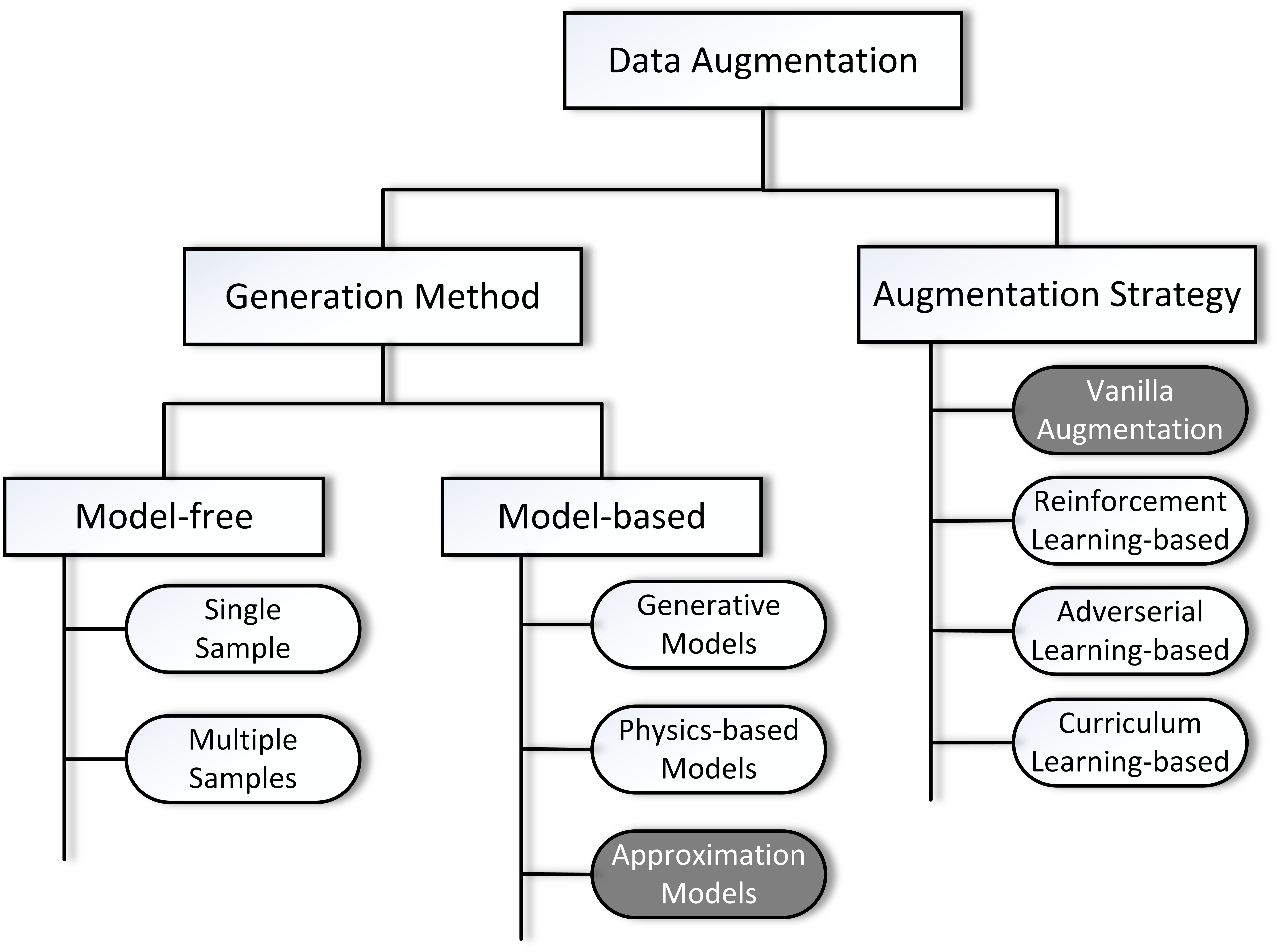}
    \caption{Methods and strategies for data augmentation.}
    \label{taxonomy}
\end{figure}

It is imperative to understand two fundamental dimensions when augmenting data into a learning model: the method used to generate the augmented data and the strategy used to train the model with the augmented data. Mingle et al. \cite{xu2022comprehensive} propose a novel augmentation taxonomy that incorporates both data generation methods and training strategies. Basically, they categorize augmentation under three titles: model-free, model-based, and optimization-based. The first two cover data augmentation methods, while the third deals with augmentation strategies.  Partially inspired by their taxonomy, our augmentation classification is illustrated in Figure~\ref{taxonomy}. 

Our approach to augmentation differs from that of \cite{xu2022comprehensive}, in which we first split the process into its two major dimensions, namely data generation methods and augmentation strategies. We divide data generation methods into two categories as model-based and model-free. Model-free data augmentation, as the name suggests, refers to techniques used to increase the diversity of training data without the use of a model\footnote{In spite of the fact that many methods, such as flipping or rotating a signal, can also be considered to include a mathematical model, they are called model-free due to their simplicity.}. Model-free data augmentation can be broadly divided into two subcategories: augmentation using a single sample or a fusion of multiple data samples. Model-free single-sample augmentation refers to the process of diversifying the dataset by applying various types of basic data processing techniques, such as rotation, scaling, flipping \cite{shorten2019survey,nalepa2019data}. Previous work such as \cite{hendrycks2019augmix} shows that when combining these techniques and combining their outputs with the original sample, further improvement is possible. 
These techniques are applied randomly, allowing for the creation of a diverse set of augmented samples from a single original source. Despite their simplicity, model-free augmentation techniques are commonly used for various computer vision tasks \cite{he2016deep,wang2021scaled,ma2020mdfn,wang2021you,liu2021swin,touvron2021training,zoph2018learning}. 

Model-free multiple-sample augmentation methods aim to combine more than one sample to dilate the training dataset distribution. Fusing two samples is the most common way of implementing multiple-sample augmentation techniques. The pioneering ``Mix-up'' \cite{zhang2017mixup} and ``Cut-Mix'' \cite{yun2019cutmix} methods are the inspiration for several subsequent augmentation methods such as \cite{walawalkar2020attentive,kim2020puzzle,verma2019manifold,uddin2020saliencymix,huang2021snapmix}.  In addition, there are several algorithms with varying strategies to utilize more than two samples for augmentation such as ``RICAP'' \cite{takahashi2018ricap} and mosaic augmentation \cite{bochkovskiy2020yolov4}. 

Model-based data augmentation refers to techniques that utilize a complex mathematical model when generating fresh samples. Under this title, we categorize the utilized models under three headings: generative models, physics-based models, and approximation models. Generative model-based augmentation methods are overwhelmingly composed of techniques that utilize Generative Adversarial Networks (GANs), Variational Autoencoders(VAEs), and neural style transfer algorithms. Following the categorization of the original GAN literature, \cite{xu2022comprehensive} further classifies generative model-based  approaches as unconditional, label-conditional, and sample-conditional. Unconditional sample generation creates fresh samples from a given input noise \cite{frid2018synthetic,madani2018chest} (also called the latent code in the GAN literature), resulting in samples that are similar to the training data distribution. Label-conditional sample generation involves conditioning the generator and discriminator models on a label \cite{condgan1,condgan2,condgan3,condgan4}, allowing for augmentation for a supervised learning model. As the final alternative, sample generation conditions can be determined by other existing samples, hence leading to label-preserving \cite{huang2018auggan,zhu2018data,geirhos2018imagenet,li2020simplified} or label-changing generation processes \cite{zhu2018emotion,zheng2021generative,schwartz2018delta,antoniou2017data,hong2021stylemix}. 

Physics-based data augmentation methods utilize physical laws and principles to create synthetic samples that are consistent with the underlying physics of the system that generates the sample space. For example, in order to augment the atmospheric turbulence effect on an image, physics-based models would rely on the actual fluid mechanics of turbulent flows in the atmosphere. The turbulence effects on 2D images are typically accurately simulated when detailed information regarding the 3D environment is provided, such as the temperature distribution and wind velocity distribution \cite{bos2012technique,schwartzman2017turbulence}. These models can provide highly detailed and accurate simulations of atmospheric turbulence effects on images, but require significant computational resources to solve the complex equations governing fluid mechanics \cite{chimitt2020simulating}. In contrast to physics-based data augmentation methods, approximation models are computationally less expensive and often do not require detailed model of the system. Instead, they utilize simplified or empirical models to approximate the generative process. Using the turbulence example, these models often do not require detailed 3D environmental data, so they are less computationally intensive. Some approximation models use statistical data that collected from different atmospheric conditions to model atmospheric effects on samples \cite{fishbain2007real,zhu2012removing}, while others use physically motivated approximations to simplify the calculation of turbulence effects on images. The simplified models aims to produce turbulent images via modelling spatially-temporal geometric distortions in pixel level \cite{mao2021accelerating,uzun2022augmentation}. However, the distinction between physics-based and approximation models can be difficult to draw in practice, regardless of the generation domain. For example, some physics-based models may use simplified systems models to reduce the computational cost, while some approximation models may still incorporate some physical principles or empirical data.

Contrary to \cite{xu2022comprehensive}, we categorize augmentation strategy as a separate dimension, because these strategies utilize data generation methods listed in the left part of our taxonomy. 
The most straightforward approach to using augmented data in a training model is to take no action and treat the enhanced samples identically to the original samples. We refer to this default approach as "Vanilla Augmentation" because this is essentially how the majority of augmentation in the deep learning literature is done. Besides the default approach, we identify three significant directions in augmentation strategies: curriculum-based, reinforcement learning-based and adversarial learning-based methods. Unlike vanilla augmentation, these titles aim at optimizing the augmentation policy. For example, curriculum learning-based augmentation strategically refines datasets in alignment with the principles of curriculum learning, ensuring a gradual and targeted enhancement process \cite{choi2024colorful}. Reinforcement learning-based data augmentation strategies aim to optimize the augmentation process based on the learning performance of the model \cite{cubuk2019autoaugment}. The philosophy behind it is to dynamically generate augmented samples that are most informative to the model and can improve its learning process. The augmentation process is modeled as a Markov Decision Process (MDP), where the state is the current training dataset, the actions are the augmentation operations, and the reward signal represents the improvement in model performance. The goal of the MDP is to find an optimal policy that generates the most informative augmented samples by balancing exploration and exploitation \cite{caicedo2015active}. Reinforcement learning-based data augmentation can lead to better performance compared to vanilla augmentation. However, it also requires more computational resources and is more complex to implement. On the other hand, adversarial-based data augmentation involves generating augmented samples that are similar to the original samples but can fool the model into making incorrect predictions. New samples are generated by seeking the small transformations of the given sample that yields maximum loss\cite{fawzi2016adaptive}. The philosophy behind it is to improve the model's robustness by exposing it to samples that are close to the original data but can cause the model to fail. The idea is that by training the model on these adversarial samples, the model will learn to be more robust and resistant to similar adversarial attacks \cite{ratner2017learning,zhang2019adversarial,peng2018jointly}. Adversarial-based data augmentation can be used in conjunction with other data augmentation methods and is particularly useful in applications where the data distribution is complex and non-linear, such as computer vision and speech recognition. However, it can also be computationally expensive and may not always lead to significant improvements in model performance. 
\subsection{Problem Definition}
The purpose of this paper is to analyze the challenges associated with augmenting for atmospheric turbulence effects in thermal adapted object detection models. By exploring this context, we aim to address the inherent complexities associated with achieving accurate and robust detection performance in real-world scenarios.

Optical systems are affected  by atmospheric turbulence, which results in noise, non-rigid distortion, and blurry images. Light rays are bent and scattered in different directions because atmospheric turbulence changes the refractive index of the air. 
What is more, turbulence in the atmosphere causes fluctuations in the air's temperature and density, which results in temporal variations in the electromagnetic radiation received by optics. For detailed explanations about the turbulence effects on optical imaging systems reader may refer to \cite{roggemann1996imaging}.

Various techniques are being studied in the literature to counteract the effects of turbulence on computer vision tasks \cite{yasarla2021learning,lou2013video,mao2022single,oreifej2012simultaneous,chen2014detecting}. Regardless of the problem, the generalization capability and robustness against challenging computer vision tasks such as detection on turbulent images, occluded images, adversarial samples etc. usually require high-quality data. In this paper, we specifically focus on thermal adapted deep object detection models, which are specialized frameworks designed to detect and localize objects using thermal imaging data. By adapting deep learning techniques to thermal imagery, these models offer enhanced detection capabilities in diverse real-world scenarios. While there has been a recent increase in research on object detection using thermal imagery \cite{wu2022uiu,li2021yolo,dai2021tirnet,jiang2022object,chen2022multi}, there are a limited number of attempts that study the atmospheric effects on thermal object detection systems \cite{oreifej2012simultaneous,uzun2022augmentation,erlenbusch2023thermal,li2023detection,xu2019robust}.

In order to counteract the effects of atmospheric turbulence, mitigation algorithms \cite{patel2022comprehensive,nieuwenhuizen2019dynamic,nair2023ddpm,jain2022evaluation} aim to reduce and/or eliminate the turbulence effects on images. Estimating and correcting turbulence distortion is a computationally intensive task that may not be feasible to perform in real-time due to limited resources. 
Another approach to overcome this difficulty, which we also explore in this paper, is to train these models with turbulent augmented samples and avoid any additional computational cost on the system.

By using turbulent image augmentation techniques, the accuracy and ability of the baseline model to handle degradation effects caused by turbulence can be improved, such as noted in \cite{uzun2022augmentation,chak2018subsampled}.Even when the test images exhibit no distortion, augmentation methods that include blurring  and geometric distortions may be useful approaches for address the increasing generalization capability of the detection models.

In this paper, we present a comparative discussion on turbulent image augmentation for thermal band images in the context of object detection. Our goal is to identify the optimal augmentation method in terms of accuracy. We generate turbulent images by employing three different atmospheric turbulence simulation methods \cite{uzun2022augmentation, chimitt2020simulating, mao2021accelerating}. The detection models used in our study include state-of-the-art real-time models, namely: RTMDet \cite{lyu2022rtmdet}, DINO \cite{zhang2022dino}, and YOLOv8 \cite{yolov8_ultralytics}.

The organization of the paper is as follows. Section 2 describes the image augmentation algorithms utilized in the paper. In Section 3, we discuss the experimental setup, specifically the datasets and techniques used to generate turbulent images. The benchmarked object detection models are presented in Section 4. Section 5 presents a discussion of the results. Finally, Section 6 concludes the paper and offers future directions.

\section{Turbulent Image Generation}
\label{sec:im_gen}
Simulation of turbulence effects on sensory signals is a major challenge in many fields, including astronomy \cite{quirrenbach2006effects}, surveillance \cite{yitzhaky2013surveillance}, and navigation \cite{stroe2016analysis}. There are various approaches to simulating these effects \cite{roggemann2018imaging}. In this paper, we utilize three \textit{approximation models} for generating turbulent images: a geometrical turbulence simulator (Section~\ref{sec:geo_sim}), a Zernike-based simulator (Section~\ref{sec:zernike_sim}), and a phase-to-space simulator (Section~\ref{sec:p2s_sim}). For the second and third simulators, camera-related parameters are required to produce physically accurate turbulent images. In order to accomplish this, we use the parameters of the FLIR Tao v2 640 x 512 13mm 45°HFoV - LWIR Thermal Imaging Camera, since "FLIR-ADASv2" dataset is collected by using this optical system. The related parameters are listed in Table~\ref{table:params_chimitt} and Table~\ref{table:mao}. The detailed explanation about the "FLIR-ADASv2" dataset and the augmented images are presented in Section~\ref{sec:dataset}.

\subsection{Geometrical Turbulence Simulator}
\label{sec:geo_sim}
Approximation models \cite{uzun2022augmentation,chak2018subsampled,mao2021accelerating} mimic the effect of atmospheric turbulence by combining blurring and random distortions. Our chosen geometric simulator \cite{uzun2022augmentation} uses a real-time model, which utilizes a geometrical turbulence approach, employing a Gaussian kernel for blurring and image warping for random distortions. In the original paper \cite{uzun2022augmentation}, the physical approximations of the geometric model are described in detail. The mathematical definition of the approach can be briefly described as Equation~\ref{eq1}.
\begin{equation}
F_n(x,y)=D((G_B(x,y) \circledast I_n(x,y)),d_n^u(x,y),d_n^v(x,y))
\label{eq1}
\end{equation}
In Equation ~\ref{eq1}, $F_n(x,y)$ is the source image, $D$ is the warping function, $\circledast$ is the convolution operation and $G_B(x,y)$ is a Gaussian kernel with variance $\sigma_B^2$, responsible for blurring. Note that warping is applied to both horizontal and vertical directions using the random distortion fields, $d_n^u(x,y)$ and $d_n^v(x,y)$, respectively, which are defined as:
\begin{equation}
d_n^u(x,y)=\gamma * (G_D(x,y) \circledast v_n^u(x,y))
\label{eq2}
\end{equation}
\begin{equation}
d_n^v(x,y)=\gamma * (G_D(x,y) \circledast v_n^v(x,y))
\label{eq3}
\end{equation}
where $\gamma$ is the amplitude of the random distortion and $G_D(x,y)$ is the Gaussian kernel with variance $\sigma_D^2$. $v_n^u(x,y)$ and $v_n^v(x,y)$ are random vectors with zero-mean, unit-variance normal distributions. Convolution operation with $G_D(x,y)$ provides spatial correlation of the random distortions over the image. $\sigma_D^2$ is used to adjust the strength of the spatial correlation while $\gamma$ is the amplitude of the distortions in the model. $\sigma_B^2$ is used to adjust the amount of blurring. Table~\ref{table:params_uzun} denotes the parameter values used for turbulent image generation.

\begin{table}[h!]
\begin{center}
\caption{Simulation parameters for the utilized geometric simulator \cite{uzun2022augmentation}.}
\begin{tabular}{ll}
\toprule
\toprule
Distortion amplitude ($\gamma$) & {[}25, 50, 100, 150{]} \\ 
{Spatial correlation strength ($\sigma_D^2$)} & 5\\ 
{Blurring ratio ($\sigma_B^2$)}  & 0.5 \\
\end{tabular}
\label{table:params_uzun}
\end{center}
\end{table}

\subsection{Zernike-based Turbulence Simulator}
\label{sec:zernike_sim}
The second model \cite{chimitt2020simulating} we used in our experiments, is another novel approximation-based approach. The authors aim at a satisfactory balance between precision and complexity for turbulence simulation. 
The model utilizes a propagation-free simulation technique to sample spatially correlated Zernike coefficients. Zernike polynomials are a set of orthogonal functions defined over a circular aperture, commonly used to represent aberrations in optical systems \cite{noll1976zernike}. These polynomials are denoted as  $\mathrm{Z}_{m}^{n}(\rho,\theta)$, as can be seen in Equation~\ref{eq:aberration}. The Zernike coefficients $\mathrm{C}_{m}^{n}$ represent the weights of each associated Zernike polynomial in a given wavefront aberration, which can be expressed as:
\begin{equation}
W(\rho,\theta) = \sum_{n= 0}^{\infty}\sum_{m = -n}^{n}\mathrm{c}_{n}^{m}\mathrm{Z}_{n}^{m}(\rho,\theta)\label{eq:aberration}
\end{equation}
In \cite{chimitt2020simulating}, Zernike coefficients are utilized to simulate atmospheric turbulence effects such as tilt and blur. The terms $\mathrm{C}_{1}^{1}$  and $\mathrm{C}_{-1}^{1}$ correspond to the horizontal and vertical tilts respectively. While these tilts signify geometric distortions across the input image, the high-order coefficients denote the blur effect.  Table~\ref{table:params_chimitt} denotes the simulation parameters used for this technique in our experiments.

\begin{table}[h!]
\begin{center}
\caption{Simulation parameters of \cite{chimitt2020simulating}.}
\begin{tabular}{ll}
\toprule
\toprule
Image Dimensions & [640, 640] pixels \\
Aperture Diameter & 29 mm \\ 
Wavelength & 10.5 $\mu$m\\ 
Refractive index parameter$(C_n^2)$  & $1e^{-15}$ \\
Focal length & 13 mm \\
Propagation Length& 2000 m \\
\end{tabular}
\label{table:params_chimitt}
\end{center}
\end{table}

\subsection{Phase-to-Space Simulator}
\label{sec:p2s_sim}

The third method we utilize for generating turbulent images is called the {Phase-to-Space (P2S) Transform} \cite{mao2021accelerating}. 
The P2S Transform, which is based on the idea of reformulating spatially varying convolution as a set of invariant convolutions using basis functions, builds upon the second method \cite{chimitt2020simulating} that we employ. The basis functions are learned by a shallow lightweight neural network. By using the learned basis functions the model can convert per-pixel Zernike coefficients to their associated point spread function (PSF) basis coefficients. This technique enables generating realistic synthetic turbulent images, while upholding theoretically verifiable statistics. Table ~\ref{table:mao} denotes the simulation parameters used for this technique in our experiments.
\begin{table}[h!]
\begin{center}
\caption{Simulation parameters of \cite{mao2021accelerating}.}
\begin{tabular}{ll}
\toprule
\toprule
Image Dimensions & [640, 640] pixels \\
Aperture Diameter & 29 mm \\ 
Wavelength & 10.5 $\mu$m\\ 
Fried parameter & 0.0145\\
Focal length & 13 mm \\
Propagation Length& 4000 m \\
\end{tabular}
\label{table:mao}
\end{center}
\end{table}
\section{Experimental Setup}
\label{sec:exp_setup}

In order to analyze the effectiveness of data augmentation over turbulent images, three state-of-the-art object detection models have been selected. A brief description of the models is provided in Section~\ref{sec:models}. To train the models, we utilize the open-source frameworks MMDetection \cite{chen2019mmdetection} and MMYOLO \cite{mmyolo2022}. The models were trained on four NVIDIA Quadro RTX 8000 GPUs. While training all of the models, the backbone weights are held frozen. Table~\ref{tab:specs_model} illustrates key specifications of three selected models, including batch size, number of epochs, augmentation techniques employed, backbone architecture, and the Average Precision (\textit{AP}) scores on the COCO dataset.
\begin{table}[H]
\caption{The configuration, and runtime settings of training processes...}
\label{tab:specs_model}
\resizebox{\textwidth}{!}{%
\begin{tabular}{c|c|c|c|c|c}
    & Batch Size & Epochs & Augmentations & Backbone & \textbf{${AP}$} on COCO\textsuperscript{1}  \\ 
    \hline
    RTMDet-x &   2 &   300 &   \begin{tabular}[c]{@{}c@{}}RandomFlip\\ RandomResize\\ RandomCrop\\ Mosaic\\ MixUp\end{tabular} &   \begin{tabular}[c]{@{}c@{}}CSPNeXt \\ with \\ P6 architecture\end{tabular} &   52.8 \\ 
    \hline
    DINO-4scale & 2 & 36 & \begin{tabular}[c]{@{}c@{}}RandomChoiceResize\\ RandomFlip\\ RandomCrop\end{tabular} & ResNet50 & 50.1 \\ 
    \hline
    YOLOv8-x & 2 & 500 & \begin{tabular}[c]{@{}c@{}}Mosaic\\ MixUpRandomFlip\\ Albumentations\end{tabular}    & YOLOv8CSPDarknet & 52.7       
\end{tabular}%
}
\noindent{\footnotesize{\textsuperscript{1} \textit{AP} scores in this table are taken from MMDetection \cite{chen2019mmdetection} and MMYOLO \cite{mmyolo2022} framework benchmark documentation, not from the original publications.}}

\end{table}

\subsection{The Dataset}
\begin{figure}[H]
        \centering
        \begin{subfigure}{0.22\textwidth}
            \includegraphics[trim=0 0 0 0,clip=true,width = 1\textwidth]{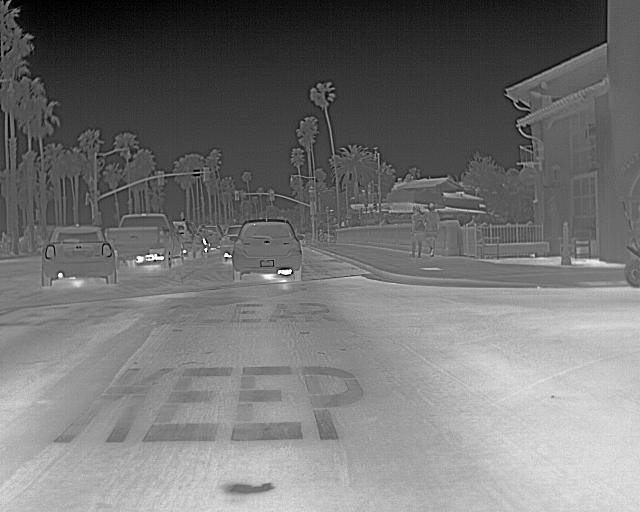}
            \includegraphics[trim=0 0 0 0,clip=true,width = 1\textwidth]{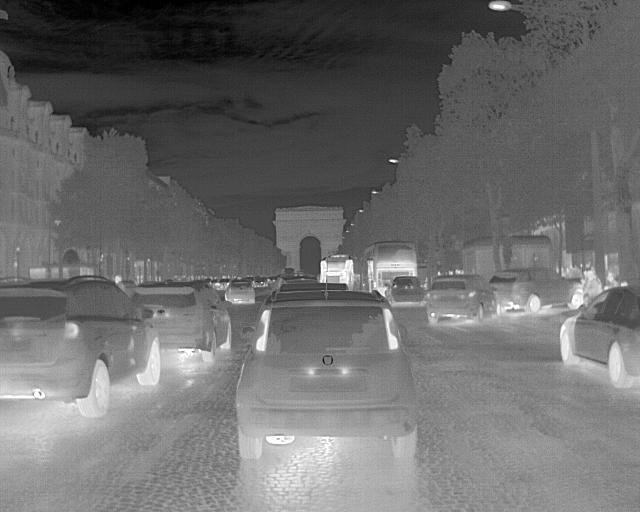}
            \includegraphics[trim=0 0 0 0,clip=true,width = 1\textwidth]{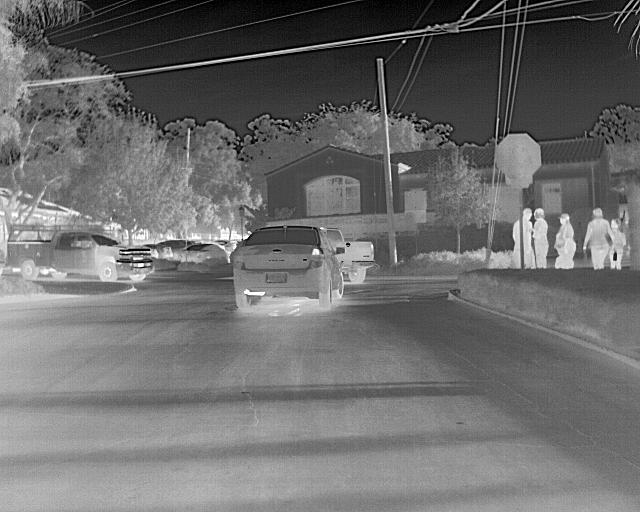}
            \includegraphics[trim=0 0 0 0,clip=true,width = 1\textwidth]{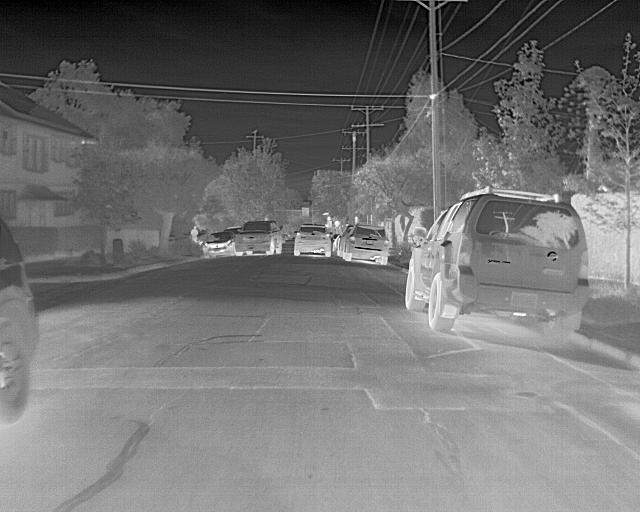}
            \includegraphics[trim=0 0 0 0,clip=true,width = 1\textwidth]{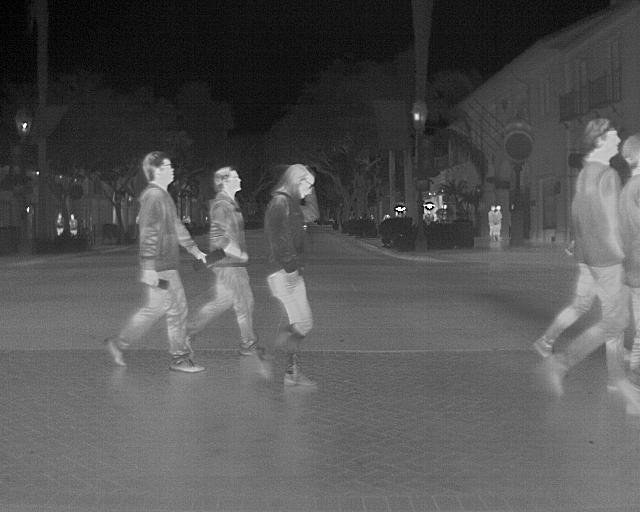}
            \includegraphics[trim=0 0 0 0,clip=true,width = 1\textwidth]{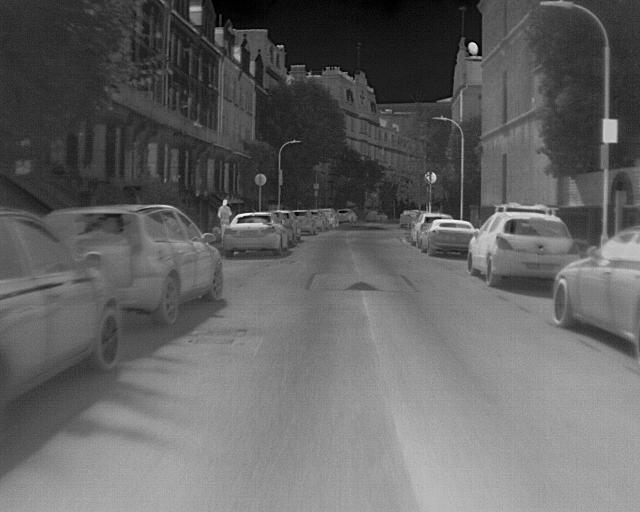}
            \caption{Original Image}
            \label{fig:clean}
        \end{subfigure}        
        \begin{subfigure}{0.22\textwidth}
            \includegraphics[trim=0 0 0 0,clip=true,width = 1\textwidth]{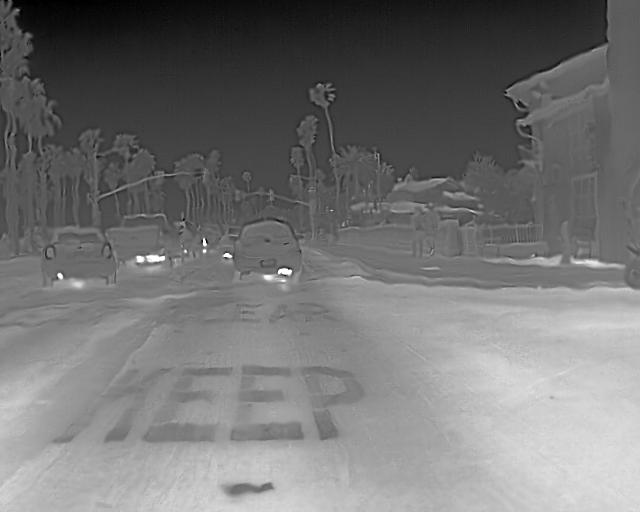}
            \includegraphics[trim=0 0 0 0,clip=true,width = 1\textwidth]{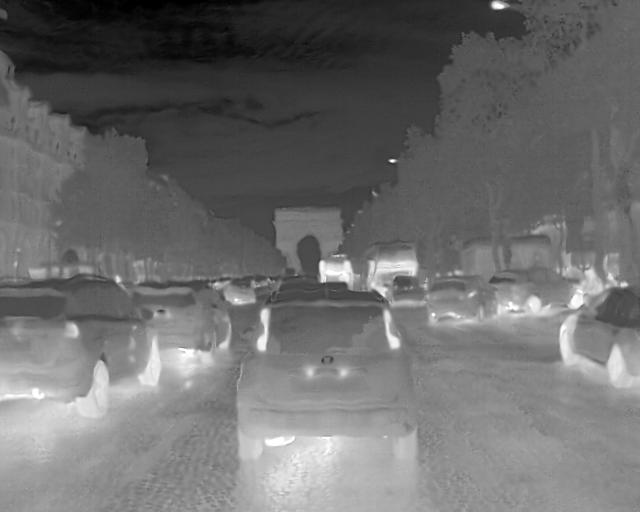}
            \includegraphics[trim=0 0 0 0,clip=true,width = 1\textwidth]{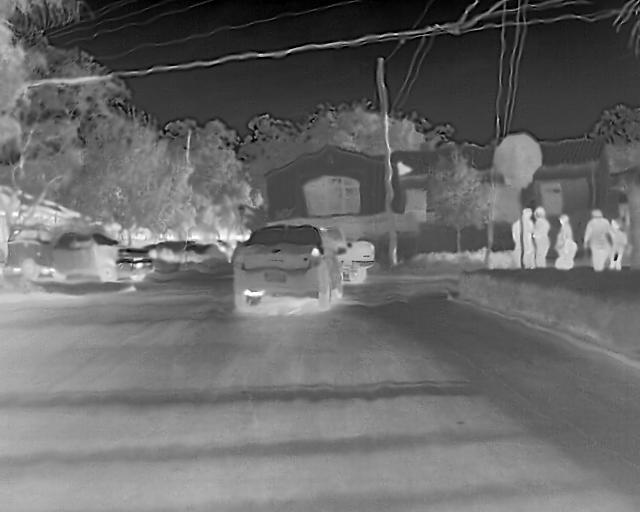}
            \includegraphics[trim=0 0 0 0,clip=true,width = 1\textwidth]{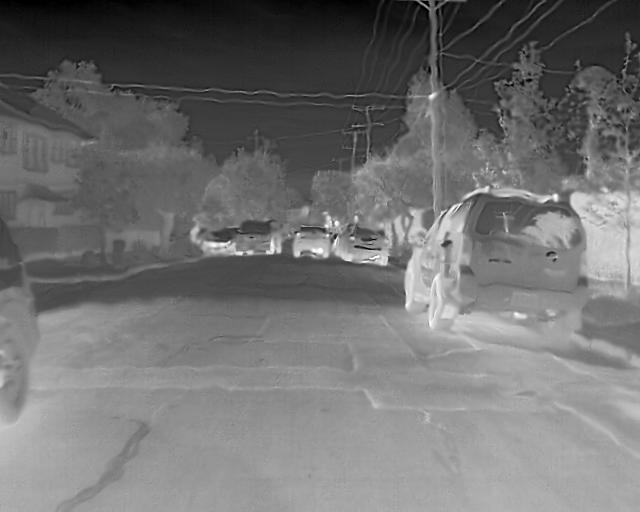}
            \includegraphics[trim=0 0 0 0,clip=true,width = 1\textwidth]{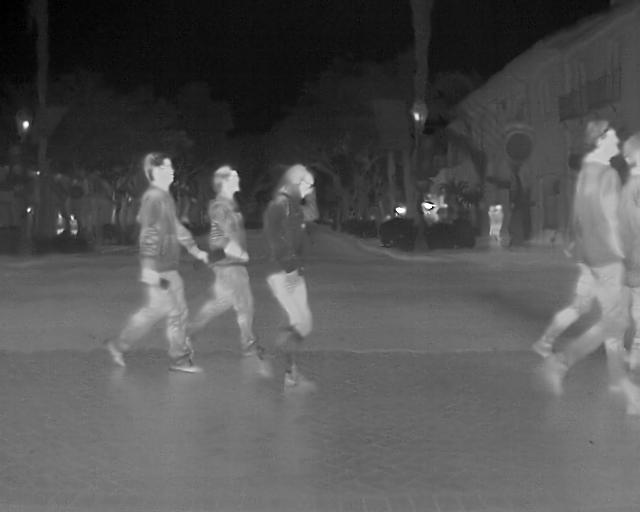}
            \includegraphics[trim=0 0 0 0,clip=true,width = 1\textwidth]{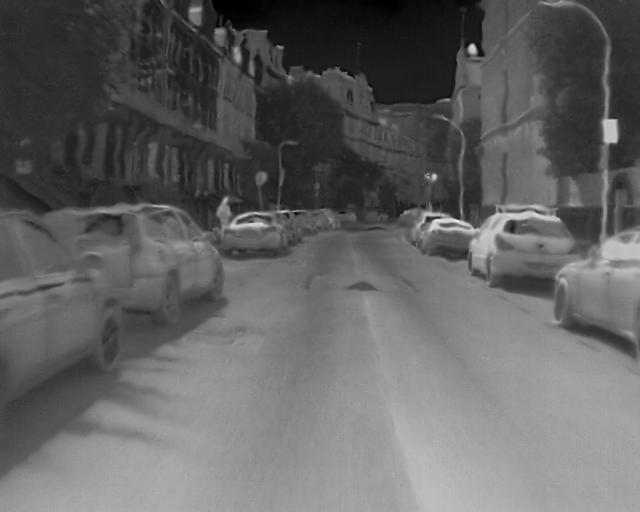}
            \caption{Geometric}
            \label{fig:gamma100}
        \end{subfigure}
        \begin{subfigure}{0.22\textwidth}
            \includegraphics[trim=0 0 0 0,clip=true,width = 1\textwidth]{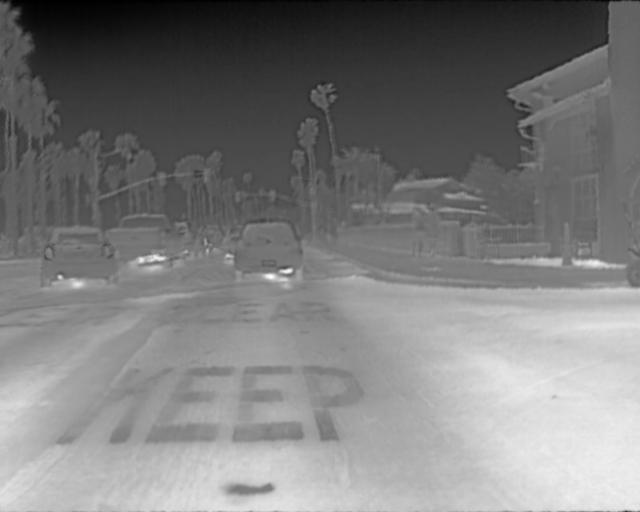}
            \includegraphics[trim=0 0 0 0,clip=true,width = 1\textwidth]{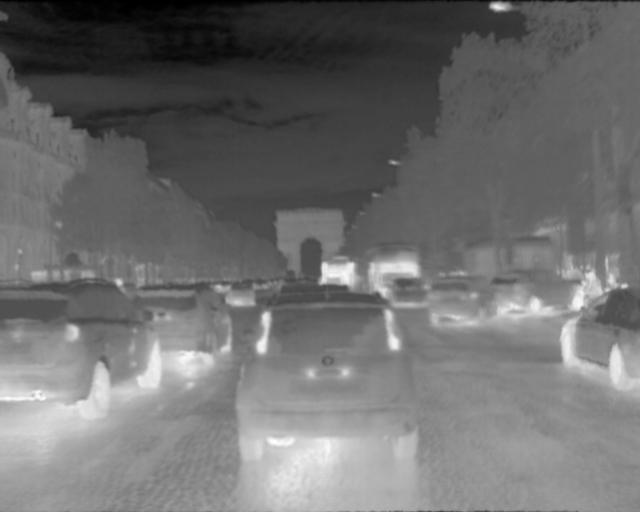}
            \includegraphics[trim=0 0 0 0,clip=true,width = 1\textwidth]{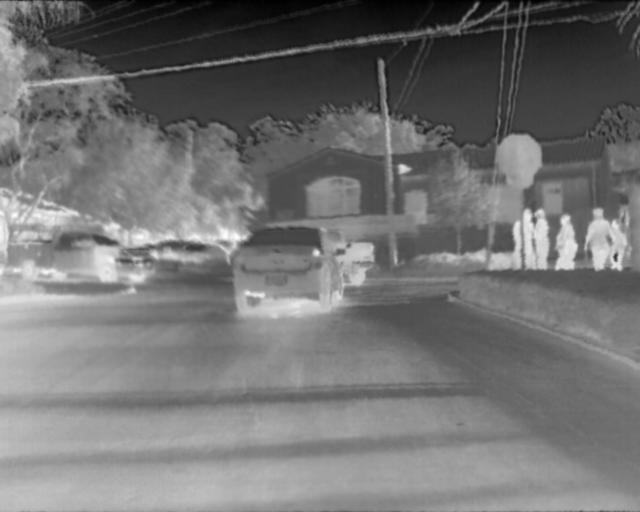}
            \includegraphics[trim=0 0 0 0,clip=true,width = 1\textwidth]{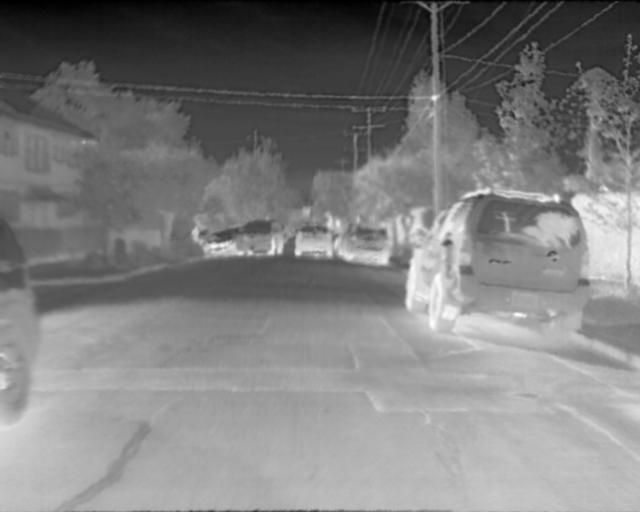}
            \includegraphics[trim=0 0 0 0,clip=true,width = 1\textwidth]{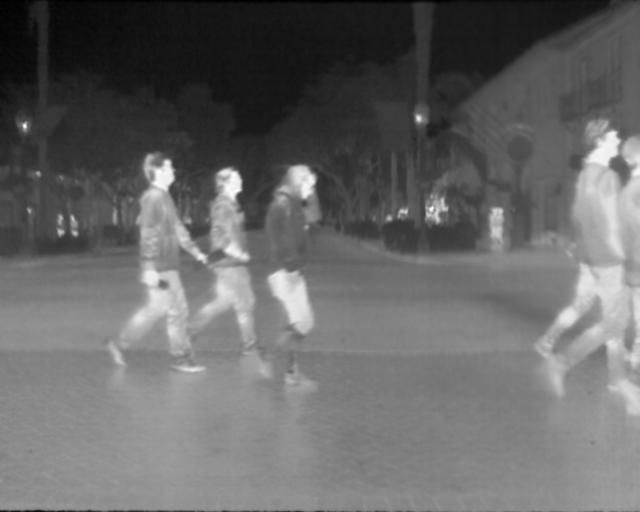}
            \includegraphics[trim=0 0 0 0,clip=true,width = 1\textwidth]{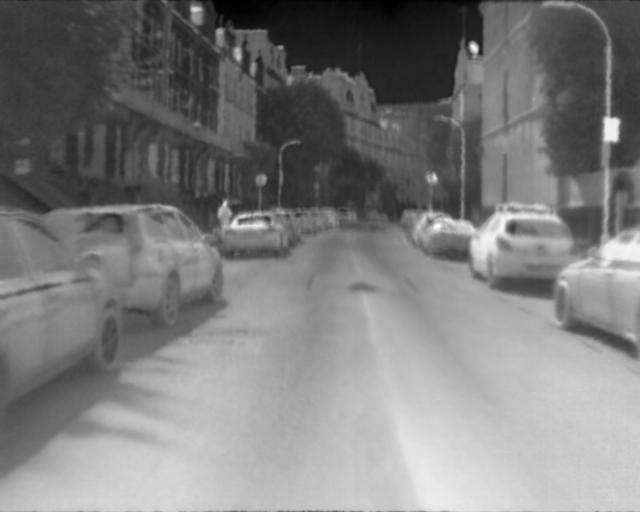}
            \caption{Zernike-based}
            \label{fig:zernike}
        \end{subfigure}
        \begin{subfigure}{0.22\textwidth}
            \includegraphics[trim=0 0 0 0,clip=true,width = 1\textwidth]{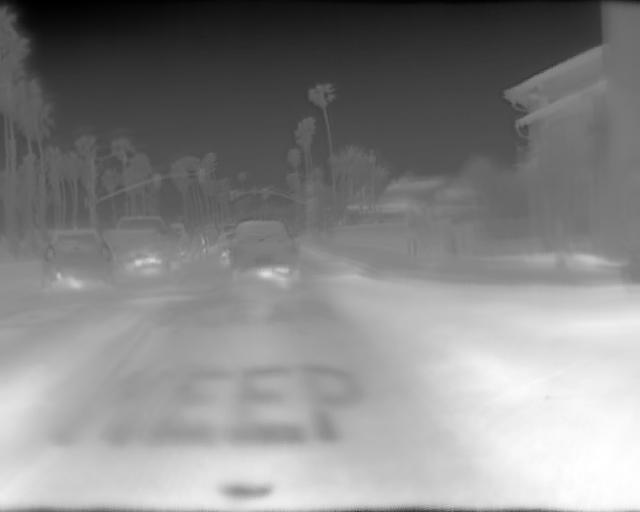}
            \includegraphics[trim=0 0 0 0,clip=true,width = 1\textwidth]{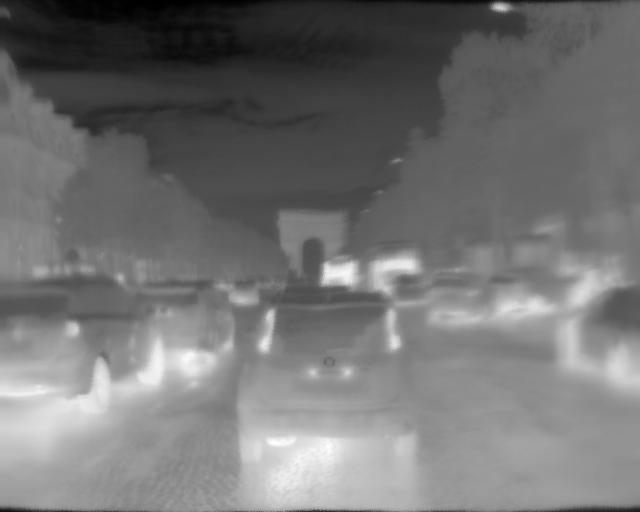}
            \includegraphics[trim=0 0 0 0,clip=true,width = 1\textwidth]{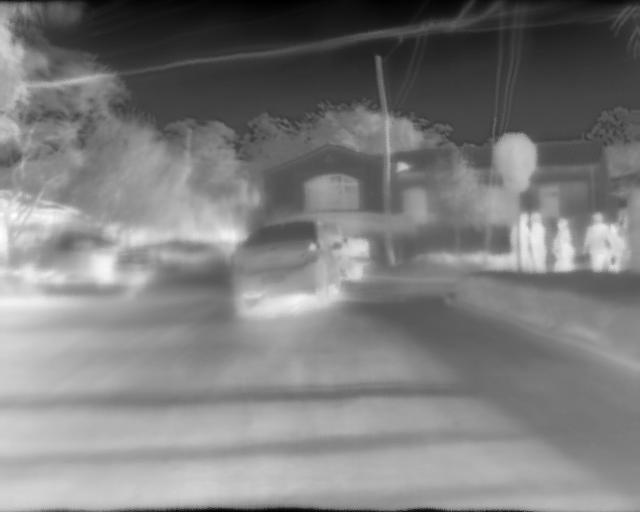}
            \includegraphics[trim=0 0 0 0,clip=true,width = 1\textwidth]{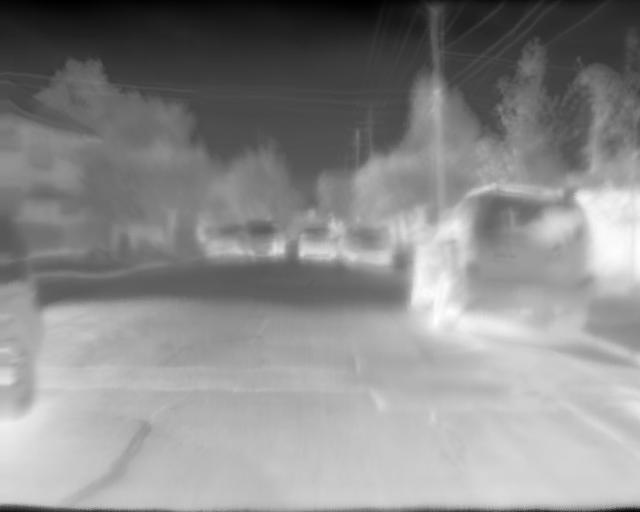}
            \includegraphics[trim=0 0 0 0,clip=true,width = 1\textwidth]{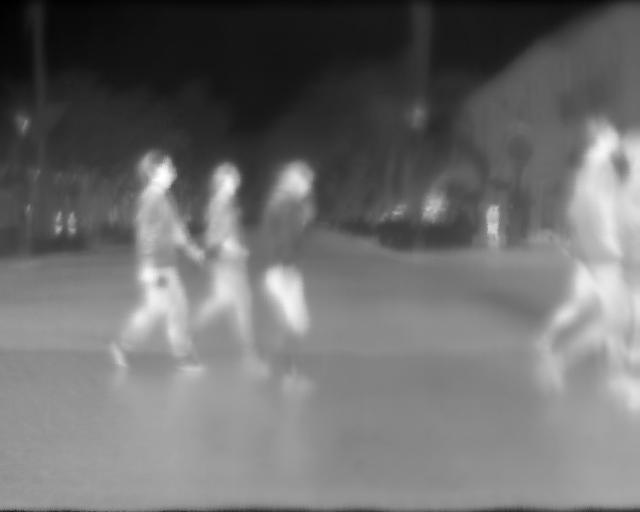}
            \includegraphics[trim=0 0 0 0,clip=true,width = 1\textwidth]{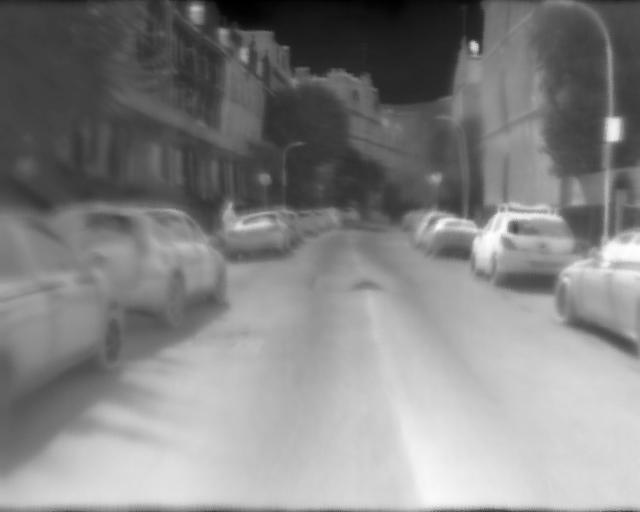}
            \caption{P2S Simulator}
            \label{fig:P2S}
        \end{subfigure}
        \caption{This figure illustrates original (input)~\ref{fig:clean} and associated outputs of three turbulence generator algorithms: geometric simulator with $\gamma = 100$~\ref{fig:gamma100}, Zernike-based simulator~\ref{fig:zernike}, and P2S Simulator~\ref{fig:P2S}.}
        \label{fig:dataset}
    \end{figure}

The experiments in this paper are carried out using the publicly available FLIR-ADAS v2 image set \cite{flir}. The experiments aim to analyze how well the augmented turbulent samples affect the performance of thermal adapted state-of-the-art object detection models. For this purpose, three different turbulence simulators are utilized to augment images. The turbulence simulators are presented in Section~\ref{sec:im_gen}. A number of original images and turbulent images generated from these originals are shown in Figure~\ref{fig:dataset}. FLIR-ADAS v2 is a medium-scale image set annotated for object detection tasks in both thermal and visible bands. The set includes a total of 26,442 annotated frames from a set of videos and still images with 15 different object classes. A total of 9,711 thermal and 9,233 RGB still images exist.  In this study, we use only the 9,711 still thermal images of the set, which are acquired with a Teledyne FLIR Tau 2. The resolution of the thermal image set is 640$\times$512 pixels. In our experiments, we focus solely on the \emph{car} and \emph{person} classes due to the insufficient number of samples available for domain adaptation in the other classes.

The original FLIR-ADAS v2 thermal still image set is divided as 90\%/10\% for training and validation respectively. Similar to previous studies that use this dataset \cite{munir2021sstn,farooq2021object}, we utilize the original 10\% validation set as our test set and create a validation set for fine-tuning purposes using the 10\% of the training set of the original FLIR-ADAS v2 thermal still image set.  

In the following, we construct the turbulent augmentation for the training using the approximation-based simulator \cite{uzun2022augmentation}, as previously mentioned above. We additionally create different versions of this turbulent training set with varying $\gamma$ levels of 25, 50, 100, and 150. We refer to this alternative set of turbulent training sets as the ``turbulent augmentation sets'', and utilize these sets as the training sets for the experiments with turbulent image augmentation.  For the test sets, we utilize all three turbulence simulators, namely geometric, Zernike-based, and P2S. For the geometric simulator, we fix the $\gamma$ value to 100 for the test samples. This way, we obtain a total of four different test sets: the original test set and the turbulent sets obtained from the three simulators.  

\label{sec:dataset}
\subsection{Object Detection Models}
In recent years, there have been remarkable advancements in deep learning-based object detection models, with innovations focusing on enhancing accuracy, efficiency, and versatility. For a survey on the subject, readers may refer to \cite{wang2023comprehensive,zou2023object}. For our experiments, we selected three state-of-the-art models, DINO \cite{zhang2022dino}, RTMDet \cite{lyu2022rtmdet}, and YOLOv8 \cite{yolov8_ultralytics}. Each of these models introduces a unique technical solution, contributing to the development of object detection literature.
\label{sec:models}
\subsubsection{DINO: DETR with Improved DeNoising Anchor Boxes for End-to-End Object Detection}
DINO \cite{zhang2022dino} represents a significant advancement in Detection Transformer (DETR) like models, with a focus on enhancing both performance and efficiency in end-to-end object detection. Central to DINO's innovation is the contrastive denoising training technique, which significantly improves training efficiency by incorporating both positive and negative samples linked to the same ground truth. This approach is further strengthened by the mixed query selection method, which enhances query initialization by merging outputs from the encoder, supplying positional queries, with learnable content queries. Additionally, DINO introduces the look forward twice scheme, a novel strategy that uses refined box information from deeper layers to fine-tune parameters in earlier layers, significantly improving accuracy. The model also integrates a deformable attention mechanism, selected for its computational effectiveness, greatly augmenting the model's attention capabilities and establishing DINO as a powerful tool in object detection.
\subsubsection{RTMDet: An Empirical Study of Designing Real-Time Object Detectors}
RTMDet \cite{lyu2022rtmdet} is a real-time object detector, designed to outperform the YOLO series and expand its capabilities into areas like instance segmentation and detection of rotated objects. One of the key innovations in RTMDet is the implementation of Large-Kernel Depth-Wise Convolutions within both its backbone and its neck. This feature enables the model to effectively capture global context, which is vital for precise object detection. 
Another notable advancement is the integration of soft labels in dynamic label assignment, a method that utilizes soft targets to enhance accuracy and diminish noise in label assignment.
\subsubsection{YOLOv8}
YOLOv8 \cite{yolov8_ultralytics}, among the pioneering object detection methods, has witnessed significant advancements with the introduction of YOLOv8, which excels in various tasks such as image classification, object detection, and instance segmentation. 
A major innovation in YOLOv8 is the introduction of the decoupled head module, which shifts from an anchor-based to an anchor-free framework. This change refines the classification and detection heads, leading to improved performance. Another key feature of YOLOv8 is its innovative approach to loss calculation, specifically the distribution focal loss in regression, customized to better suit the model. Additionally, YOLOv8 utilizes advanced data augmentation techniques, particularly by disabling mosaic augmentation towards the end of its training. This enhancement optimizes the training process and significantly improves the model's accuracy and efficiency across various applications.

\section{Results}
\label{sec:results}
In this section, we present the results of our experiments. The evaluation metrics are based on the COCO (Common Objects in Context) metrics \cite{lin2014microsoft}. The evaluated scores of the metrics are presented as  Tables~\ref{tab:resultsTable},~\ref{tab:aug_resultsTable},~\ref{tab:zernike_results},~\ref{tab:P2S_results}. The tables have identical layout. The first column indicates whether turbulent image augmentation was applied during training (w/ for "with" and w/o for "without").  The second column lists the model names. The next six columns represent varying Average Precision $(AP)$ metrics: $AP$, $AP_{50}$, $AP_{75}$, $AP_{S}$, $AP_{M}$, and $AP_{L}$. $(AP)$ values are computed by calculating the area under the precision-recall curve for IoU thresholds ranging from 0.50 to 0.95. Similarly, mean AP values are determined specifically at IoU thresholds of 0.50 and 0.75 by averaging the AP values obtained from the precision-recall curve at these thresholds, respectively. Each row corresponds to a specific experiment, showing the performance metrics of the corresponding model under the given condition. Horizontal lines separate different sections of the table, providing clarity in distinguishing between experiments conducted with different models. Tables ~\ref{tab:resultsTable},~\ref{tab:aug_resultsTable},~\ref{tab:zernike_results}, and ~\ref{tab:P2S_results} respectively present the results obtained using the original (i.e. clean) test dataset, geometric turbulence simulator, Zernike-based turbulence simulator, and P2S turbulence simulator applied to the test dataset.

\begin{table}[H]
\caption{Mean Average Precision results obtained for different experiments using each detection model, with or without turbulent image augmentation while training for varying levels of turbulence gain $\gamma$. Only the original test is utilized in these experiments.}

\newcolumntype{C}{>{\centering\arraybackslash}X}
\begin{tabularx}{\textwidth}{ClCCCCCC}

\multicolumn{8}{c}{ \textbf{Clean Test Set.}}  \\
\\
\textbf{Aug} & \textbf{Model} & \textit{$\textbf{AP}$} & \textit{$\textbf{AP}_{\textbf{50}}$} & \textit{$\textbf{AP}_{\textbf{75}}$} & \textit{$\textbf{AP}_{\textbf{S}}$} & \textit{$\textbf{AP}_{\textbf{M}}$}  & \textit{$\textbf{AP}_{\textbf{L}}$}  \\ 
\midrule
\midrule
w/o   & RTMDet-x & 56.8 & 84.3 & 60.5 & 45.3 & 78.3 & 81.8 \\ 
w/  & RTMDet-x   & 58.2 & 85.2 & 63.1 & 47.2 & 79.1 & 82.3 \\ 
\hline

w/o  & DINO-4scale & 56.3 & 85.9 & 59.9 & 45.7 & 75.5 & 81.1 \\
w/   & DINO-4scale & 57.8 & 87.7 & 61.8 & 47.5 & 77.1 & 81.8 \\ \hline
w/o   & YOLOv8-x   & 57.8 & 85.0 & 62.4 & 47.0 & 78.9 & 80.0 \\ 
w/  & YOLOv8-x     & 58.4 & 85.6 & 62.9 & 47.6 & 79.2 & 81.5 \\
\hline
\end{tabularx}%
\label{tab:resultsTable}
\end{table}
\begin{table}[H]
\caption{Mean Average Precision results obtained for different experiments using each detection model, trained with and without turbulent augmentation set, and for varying levels of turbulence gain $\gamma$. For this experiment, the test set is constructed using the geometric simulator \cite{uzun2022augmentation}.}

\newcolumntype{C}{>{\centering\arraybackslash}X}
\begin{tabularx}{\textwidth}{ClCCCCCC}

\multicolumn{8}{c}{ \textbf{Test set with turbulence $\gamma$ = 100}}  \\
\\
\textbf{Aug} & \textbf{Model} & \textit{$\textbf{AP}$} & \textit{$\textbf{AP}_{\textbf{50}}$} & \textit{$\textbf{AP}_{\textbf{75}}$} & \textit{$\textbf{AP}_{\textbf{S}}$} & \textit{$\textbf{AP}_{\textbf{M}}$}  & \textit{$\textbf{AP}_{\textbf{L}}$}  \\ 
\midrule
\midrule
w/o & RTMDet-x & 41.8 & 69.9 & 41.9 & 27.8 & 67.7 & 78.2 \\
w/ & RTMDet-x  & 47.1 & 76.4 & 47.9 & 33.3 & 72.7 & 81.7 \\ \hline
w/o & DINO-4scale & 35.9 & 65.9 & 34.1 & 23.0 & 59.4 & 75.2 \\
w/ & DINO-4scale  & 44.6 & 77.2 & 43.9 & 32.0 & 67.7 & 78.5 \\ \hline
w/o & YOLOv8-x    & 43.1 & 72.5 & 42.8 & 29.5 & 68.8 & 77.4 \\
w/ & YOLOv8-x     & 47.8 & 77.6 & 48.6 & 34.6 & 73.1 & 79.1 \\ 
\hline

\end{tabularx}%
\label{tab:aug_resultsTable}
\end{table}
\begin{table}[H]
\caption{Mean Average Precision results obtained for different experiments using each detection model, trained with and without turbulent augmentation set, and for varying levels of turbulence gain $\gamma$. For this experiment, the test set is constructed using the Zernike-based simulator \cite{chimitt2020simulating}.}
\newcolumntype{C}{>{\centering\arraybackslash}X}
\begin{tabularx}{\textwidth}{ClCCCCCCC}

\multicolumn{8}{c}{ \textbf{Test set Zernike-method.}}  \\
\\
\textbf{Aug} & \textbf{Model} & \textit{$\textbf{AP}$} & \textit{$\textbf{AP}_{\textbf{50}}$} & \textit{$\textbf{AP}_{\textbf{75}}$} & \textit{$\textbf{AP}_{\textbf{S}}$} & \textit{$\textbf{AP}_{\textbf{M}}$}  & \textit{$\textbf{AP}_{\textbf{L}}$}  \\ 
\midrule
\midrule
w/o & RTMDet-x & 36.2 & 63.5 & 34.9 & 20.7 & 65.9 & 76.0 \\
w/  & RTMDet-x & 40.8 & 67.0 & 37.2 & 24.8 & 67.7 & 79.9 \\ \hline
w/o & DINO-4scale & 31.7 & 60.3 & 28.7 & 17.4 & 57.2 & 72.0 \\
w/  & DINO-4scale & 34.9 & 64.3 & 32.1 & 20.3 & 61.7 & 75.1  \\ \hline
w/o & YOLOv8-x    & 36.6 & 64.9 & 35.1 & 21.5 & 66.5 & 77.3 \\
w/  & YOLOv8-x    & 41.2 & 68.2 & 37.5 & 24.9 & 68.8 & 79.7 \\

\hline
\end{tabularx}%
\label{tab:zernike_results}
\end{table}
\begin{table}[H]
\caption{Mean Average Precision results obtained for different experiments using each detection model, trained with and without turbulent augmentation set, and for varying levels of turbulence gain $\gamma$. For this experiment, the test set is constructed using the P2S-based simulator\cite{mao2021accelerating}.}

\newcolumntype{C}{>{\centering\arraybackslash}X}
\begin{tabularx}{\columnwidth}{ClCCCCCC}

\multicolumn{8}{c}{ \textbf{Test set with turbulence based-on P2S-method}}  \\
\\
\textbf{Aug} & \textbf{Model} & \textit{$\textbf{AP}$} & \textit{$\textbf{AP}_{\textbf{50}}$} & \textit{$\textbf{AP}_{\textbf{75}}$} & \textit{$\textbf{AP}_{\textbf{S}}$} & \textit{$\textbf{AP}_{\textbf{M}}$}  & \textit{$\textbf{AP}_{\textbf{L}}$}  \\ 
\midrule
\midrule
w/o & RTMDet-x  & 19.7 & 38.3 & 17.5 & 6.4 & 42.2 & 67.5 \\
w/  & RTMDet-x  & 24.5 & 49.1 & 21.2 & 9.6 & 49.8 & 74.5 \\ \hline
w/o & DINO-4scale  & 17.1 & 35.9 & 14.4 & 5.3 & 37.0 & 64.7 \\
w/  & DINO-4scale  & 19.5 & 42.5 & 16.6 & 7.5 & 39.6 & 66.9  \\ \hline
w/o & YOLOv8-x     & 23.0 & 46.7 & 19.5 & 7.6 & 47.7 & 69.2 \\
w/  & YOLOv8-x     & 24.8 & 49.9 & 21.4 & 9.8 & 49.9 & 73.0 \\ 
\hline
\end{tabularx}%
\label{tab:P2S_results}
\end{table}

\section{Discussion}

The experimental results reported in Tables ~\ref{tab:resultsTable}, ~\ref{tab:aug_resultsTable},~\ref{tab:zernike_results}, and ~\ref{tab:P2S_results} offer valuable insights into the efficiency of turbulence image augmentation for improving object detection performance under challenging atmospheric turbulence conditions. Below is a list of salient observations based on these results.

Firstly, as illustrated in Table~\ref{tab:resultsTable}, utilizing turbulence image augmentation during the training phase leads to improved detection performance independent from the evaluated models, even when testing on the clean test set. This improvement suggests that augmenting with turbulent samples can help the models generalize better and become more robust to various image degradations caused by atmospheric turbulence.

Furthermore, the results presented in Tables ~\ref{tab:aug_resultsTable},~\ref{tab:zernike_results}, and ~\ref{tab:P2S_results} clearly present the critical importance of turbulence augmentation when the set comprises images degraded by atmospheric turbulence. Across all three turbulence simulators employed, the models trained with turbulence augmentation consistently outperform their non-augmented counterparts. This pattern underlines the necessity of including turbulence specific augmentations to enhance detection accuracy in real-world scenarios affected by atmospheric distortions. 

We observe that there is a noticeable performance improvement associated with turbulence augmentation for smaller objects (see $AP_{S}$ in tables) compared to larger objects(see $AP_{L}$ in tables). This observation aligns with the expected impact of atmospheric turbulence, which is clearly observable in Fig~\ref{fig:dataset}. Atmospheric turbulence tends to have a more significant disruptive effect on the visibility and detection of smaller objects due to their limited spatial area and lower contrast. 

While all evaluated models benefit from turbulence augmentation, the degree of improvement varies across different architectures. For instance, the RTMDet-x and YOLOv8-x models seem to indicate more significant performance enhancement compared to DINO-4scale, as evidenced by the larger differences in AP scores between their augmented and non-augmented versions that shown in Tables~\ref{tab:aug_resultsTable},~\ref{tab:zernike_results}, and ~\ref{tab:P2S_results}.

Table~\ref{tab:P2S_results} illustrates the most drastic simulation results compared to the other two simulators. Performance degradation is more pronounced when P2S-simulated tests are evaluated. This observation and Figure~\ref{fig:dataset} suggest that the P2S simulator may produce more challenging turbulent images. Due to employed harsh parameter setting in this study, given in Table~\ref{table:mao}.


\section{Conclusions}
The present study explores the effectiveness of turbulence image augmentation for improving the performance of thermal-adapted object detection models in the presence of atmospheric turbulence. The results demonstrate that utilizing turbulence-specific augmentations during training can significantly improve detection accuracy and robustness against turbulence degradations. By employing three distinct approximation-based turbulence simulators (geometric, Zernike-based, and P2S) and evaluating on corresponding turbulent test sets, this study provides an extensive analysis of the impact of turbulence augmentation across various simulation approaches. The findings indicate that turbulence augmentation is beneficial for all evaluated models, including RTMDet-x, DINO-4scale, and YOLOv8-x. However, the degree of improvement varies across architectures, with some models demonstrated significant improvements than others. This observation suggests that the effectiveness of turbulence augmentation may be influenced by the specific design and characteristics of the object detection model.

Furthermore, the results suggest that turbulence augmentation is particularly effective in improving the detection of smaller objects, which are more affected to the effects of atmospheric turbulence due to their limited spatial area and lower contrast. This finding aligns with the expectations and highlights the importance of addressing the challenges by atmospheric turbulence for detection of small-scale objects.

While the current study focused on thermal-adapted object detection models, the insights gained from this work may be applicable to other computer vision tasks and modalities affected by atmospheric turbulence, such as visible-light object detection, tracking, or segmentation. Future research could explore the generalization of these findings to different domains and applications.

Potential future directions for this line of research could include investigating more advanced turbulence simulation techniques, such as physics-based models or data-driven approaches, to better capture the complexities of real-world atmospheric conditions. Additionally, exploring more sophisticated augmentation strategies, such as curriculum learning or reinforcement learning-based methods, may further improve the effectiveness of turbulence augmentation by optimizing the augmentation process. 

Overall, this study provides valuable insights for researchers and engineers working on computer vision systems operating in challenging atmospheric conditions, highlighting the importance of utilizing turbulence-specific augmentations to improve detection accuracy and robustness. The findings contribute to the ongoing efforts in developing robust and reliable computer vision solutions for real-world applications affected by atmospheric turbulence.

\bibliographystyle{unsrt}
\bibliography{bibliography}



%


\end{document}